\documentclass[12pt]{article}
\RequirePackage{amsthm,amsmath,amsfonts,amssymb}
\usepackage{natbib}
\RequirePackage[colorlinks,citecolor=blue,urlcolor=blue]{hyperref}
\RequirePackage{graphicx}

\newcommand{\Rcal}{\mathcal R}

\newcommand{\Tcal}{\mathcal T}

\usepackage{comment}
\usepackage{amsmath}
\usepackage{graphicx,psfrag,epsf}
\usepackage{enumerate}
\usepackage{natbib}
\usepackage{url} 

\usepackage[ruled,linesnumbered]{algorithm2e}
\usepackage{pifont}
\usepackage{soul}
\usepackage{booktabs}
\usepackage{multirow}
\usepackage{bigdelim}
\usepackage{sistyle}
\usepackage{graphicx}
\usepackage{epstopdf}
\usepackage{mathrsfs}
\usepackage{nicematrix}
\usepackage{caption}
\usepackage{subcaption} 
\usepackage{multirow}
\usepackage{color}
\usepackage{xcolor}
\usepackage{tabularx}
\usepackage{url}
\usepackage{svg}
\usepackage{graphicx} 
\usepackage{subcaption} 

\usepackage{authblk}

\theoremstyle{plain}

\usepackage[T1]{fontenc}

\begin{document}

\def\spacingset#1{\renewcommand{\baselinestretch}%
{#1}\small\normalsize} \spacingset{1}

\title{\bf Spatio-temporal Prediction of Fine-Grained Origin-Destination Matrices with Applications in Ridesharing}
\author[]{} 
\author{Run Yang$^1$, Runpeng Dai$^2$, Siran Gao, Xiaocheng Tang$^3$, Fan Zhou$^{1*}$, and Hongtu Zhu$^{2}$\thanks{Correspondence to: Fan Zhou <zhoufan@mail.shufe.edu.cn>, Shanghai University of Finance and Economics, and Hongtu Zhu <htzhu@email.unc.edu>, University of North Carolina at Chapel Hill }\\ 
$^1$Department of Statistics and Management \\
    Shanghai University of Finance and Economics \\
    $^2$Department of Biostatistics
    \\ 
    University of North Carolina at Chapel Hill\\ 
    $^3$Personal Researcher, California, USA
} 
\date{} 
\maketitle

\bigskip
\begin{abstract}
Accurate spatial-temporal prediction of network-based travelers' requests is crucial for the effective policy design of ridesharing platforms. Having knowledge of the total demand between various locations in the upcoming time slots enables platforms to proactively prepare adequate supplies, thereby  
increasing the likelihood of fulfilling travelers' requests and redistributing idle drivers to areas with high potential demand to optimize the global supply-demand equilibrium. This paper delves into the prediction of Origin-Destination (OD) demands at a fine-grained spatial level, especially when confronted with an expansive set of local regions. While this task holds immense practical value, it remains relatively unexplored within the research community. To fill this gap, we introduce a novel prediction model called OD-CED, which comprises an unsupervised space coarsening technique to alleviate data sparsity and an encoder-decoder architecture to capture both semantic and geographic dependencies. Through practical experimentation, OD-CED has demonstrated remarkable results. It achieved an impressive reduction of up to 45\% reduction in root-mean-square error and 60\% in weighted mean absolute percentage error over traditional statistical methods when dealing with OD matrices exhibiting a sparsity exceeding 90\%.

\end{abstract}

\noindent%
\vfill

\newpage
\spacingset{1.9} 
\section{Introduction}
\label{sec:intro}

Spatial-temporal processes have gained prominence in contemporary statistics and data science, driven by its extensive applications across fields like climate modeling,   traffic management, public health surveillance \citep{wikle2023statistical}. For instance, dynamic Origin-Destination (OD) matrices are pivotal in quantifying the flow between various spatial entities over time. However, analyzing and predicting OD matrices at a fine-grained spatial scale involving numerous locations present significant challenges within the research community  \citep{zhang2021dneat, wang2019origin, ke2021predicting}. Ride-sourcing platforms such as Uber exemplify this complexity.  When a city area is divided into $N$ locations, the OD demand flows during each time interval across these locations generate an $N \times N$ matrix. This matrix illustrates the travel requests between any pair of spatial cells, offering a fine-grained view of traffic movements. Such detailed OD matrices are crucial for developing more effective transportation strategies, as they provide comprehensive micro-level traffic insights, enabling precise traffic condition forecasts. However, the task of predicting large-scale, fine-grained OD matrices introduces three key challenges:

\textbf{Scalability.} The OD matrix grows dramatically with the number of spatial divisions $N$. For instance, analyzing traffic data across hundreds of locations can yield an OD matrix with hundreds of thousands of flows. Addressing these large matrices requires scalable prediction models, a challenge that many advanced methods struggle with. As spatial complexity increases, scalability issues arise, including rising memory requirements and decreased computational efficiency.

\textbf{Data Sparsity.} 
Adopting a fine-grained spatial approach often leads to a decrease in average demands at both individual locations and across flows, thereby heightening the issue of data sparsity.  In the ridesharing industry, segmented urban areas frequently exhibit near-zero customer requests between distant or unrelated locations for the majority of the time, resulting in skewed OD demand distributions. This accentuates the challenges posed by irregular traffic patterns and lessens the effectiveness of identifying and leveraging patterns or group structures within OD data.

\textbf{Semantic and Geographical Dependencies.} Both semantic and geographical dependencies play crucial roles in OD prediction. Higher semantic dependence often leads to larger travel demands. For instance, two locations such as a residential area and a business district may be semantically correlated, even if they are geographically distant. Conversely, geographically adjacent cells are more likely to share similar traffic patterns due to similarities in their points of interest (POIs), even if the semantic dependency is weak in this scenario.  However many current prediction methods overlook these two dependencies, which can make their forecasts less accurate.

\subsection{Related work}
OD data has been extensively studied by statisticians \citep{medina2002traffic, hazelton2008statistical, ma2018statistical}, focusing on estimating OD matrices or understanding commuting and migration patterns. We categorize these methods based on the data used.

One category employs data aggregation methods, which combines diverse sources like Call Detail Records (CDR) \citep{calabrese2011estimating}, mobile cellular signaling data \citep{janzen2016estimating}, and check-in data \citep{fekih2021data}. Another category uses models like the gravity \citep{pourebrahim2018enhancing} and radiation \citep{liu2020urban} models to link OD matrices with factors like population, socioeconomic variables, land types, and distances.

Much research focuses on OD estimation, often using Bayesian modeling \cite{vardi1996network}. Other methods include Bayesian inference \citep{tebaldi1998bayesian}, information-theoretic approaches \citep{zhang2005estimating}, and feature reduction models \citep{lakhina2004structural}, applied in fields like internet anomaly detection \citep{papagiannaki2003long}.

This paper addresses OD forecasting, predicting future OD matrices based on historical data. Traditional methods focus on OD estimation or matrix completion and are not suited for forecasting. Models like ARIMA and Support Vector Regression (SVR) are limited due to inefficiency and lack of spatial correlation capture. Advanced methodologies are needed for efficient predictions capturing spatial dependencies.

Existing computational OD forecasting methods generally fall into two distinct categories. The first category conceptualizes the OD matrix as a single image and utilizes Convolutional Neural Networks (CNNs) for capturing spatial-temporal closeness, as seen in references \citep{he2016deep,zhang2016dnn,zhang2017deep, ma2017learning,liu2019contextualized}.  A notable advancement in this category is the integration of Recurrent Neural Networks (RNNs) with CNNs, enabling more effective capture of long-term spatial-temporal correlations. ConvLSTM \citep{shi2015convolutional}, initially developed for weather forecasting, has notably achieved success in traffic flow predictions. Further improvements include the integration of non-local modules with ConvLSTM \citep{liu2019contextualized} and the introduction of a look-up convolution layer, termed LC, for enhanced learning of time-series traffic patterns \citep{lv2018lc}.

The second category, exemplified by \citep{wang2019origin, zhang2021dneat}, approaches the OD matrix as a fully connected weighted bidirectional graph, employing Graph Neural Networks (GNNs) to discern dynamic traffic patterns. Innovations in this category include combining Graph Convolutional Networks (GCNs) and CNNs to capture spatio-temporal interactions, as in STGCN \citep{lu2020spatiotemporal}, and the development of multi-GCN architectures \citep{chai2018bike, geng2019spatiotemporal, lu2020spatiotemporal}. Techniques such as ASTGCN \citep{guo2019attention} and STGNN \citep{wang2020traffic} incorporate spatio-temporal attention for improved GCN embedding, while SFGNN \citep{song2020spatial} forms a spatio-temporal fusion graph. Other notable approaches include the use of self-adaptive adjacency matrices \citep{wu2019graph}, Adaptive Graph Convolutional Recurrent Networks \citep{bai2020adaptive}, and the STSGCN \citep{mengzhang2020spatial} for learning dynamic spatial-temporal correlations. However, a significant limitation of these methods, across both categories, is the dramatic increase in spatiotemporal complexity as $N$ grows, posing scalability challenges in effectively handling large-scale OD matrices.

\subsection{Contributions}

To tackle the aforementioned challenges, we introduce a novel lightweight OD prediction model called the \textbf{{C}oarseing}-\textbf{\underline{E}ncoder}-\textbf{\underline{D}ecoder} network for fine-grained \textbf{\underline{O}rigin}-\textbf{\underline{D}estination} data (\textbf{OD-CED}). As shown in Fig ~\ref{framework}, the OD-CED framework comprises two primary stages: 

$\bullet$ \textbf{Preprocess Stage.} We transform $N$ fine-grained cells into $M$ coarse-grained super-cells to significantly reduce computational demands ($M \ll N$) while preserving the essential spatial-temporal dependencies. Through a strategic two-step Label Propagation (LP) process, cells exhibiting similar OD patterns are merged into larger clusters, thereby enhancing the framework’s robustness against irregular traffic demands.

$\bullet$ \textbf{Learning Stage.}
We utilize a hierarchical embedding architecture to capture semantic and geographical dependencies. An encoder learns the representations of super-cells through a multi-head self-attention network and the decoder predicts fine-grained cells by estimating the similarity between each fine-grained cell and its corresponding super-cells. 

In particular, the new method tries to address three  critical questions:
(Q1) How can we reduce prediction errors for OD flows with typically non-zero values, given that the model often yields sparse predictions due to the prevalence of zero values across most spatio-temporal locations?
(Q2) How can we accurately predict exceptional or extreme non-zero demands, which may arise when fine-grained spatial divisions result in less evident periodic patterns?
(Q3) How can we utilize both semantic and geographic information to better understand traffic patterns of OD demands?
These questions drive the methodological development outlined in Section \ref{sec3}.

The main contributions of this paper are as follows:
(i) We are the first to address large-scale fine-grained OD prediction with extremely sparse data.
(ii) We propose a novel method to merge fine-grained cells into super-cells, reducing OD matrix size while maintaining dependencies.
(iii) We design an encoder-decoder model to capture global dependencies effectively.
(iv) We develop a permutation-invariant module to learn super-cell level representations.

\section{Fine-grained origin-destination data in ridesharing}\label{sec2}

In this section, we analyze two real industrial datasets collected from one ridesharing company to illustrate the challenges caused by fine-grained OD. In both datasets, each day was divided into 24 equally spaced non-overlapping time intervals (1 hou) and the entire city area was partitioned into $N$ local regions. The number of customers' travel requests starting from any origin place and ending at any other destination place within each time interval was recorded. In this case, each OD flow represents the total demand between two spatial locations for each single hour.

Below is a brief introduction of the two datasets.

\textbf{City-C:} City-C is an anonymized dataset processed and released by the ridesharing company. It comprises historical ride-hailing orders spanning from November $1$st to November $30$th, $2016$ in City C.
The urban area is segmented into $632$ hexagonal cells, each covering an area of around $1.2$ square kilometers. Hourly counts of ride-hailing orders between any two cells are recorded.
The sparsity ratio of City-C is \textbf{$99.15\%$}.

\textbf{City-S:}. City-S is a real internal dataset collected from June $1$st to November $30$th, 2021 in City S, encompassing a longer time duration (180 days) compared to City-C (30 days). For City-S, each partitioned local region spans $2.1km^{2}$, with a total of $638$ such segments covering the entire city area. The sparsity ratio of City-S is \textbf{$98.02\%$}, slightly lower than that of City-C, suggesting that City-S experiences higher overall travel demands.

\begin{figure*}[!hbt]
\centering
\includegraphics[width=\textwidth]{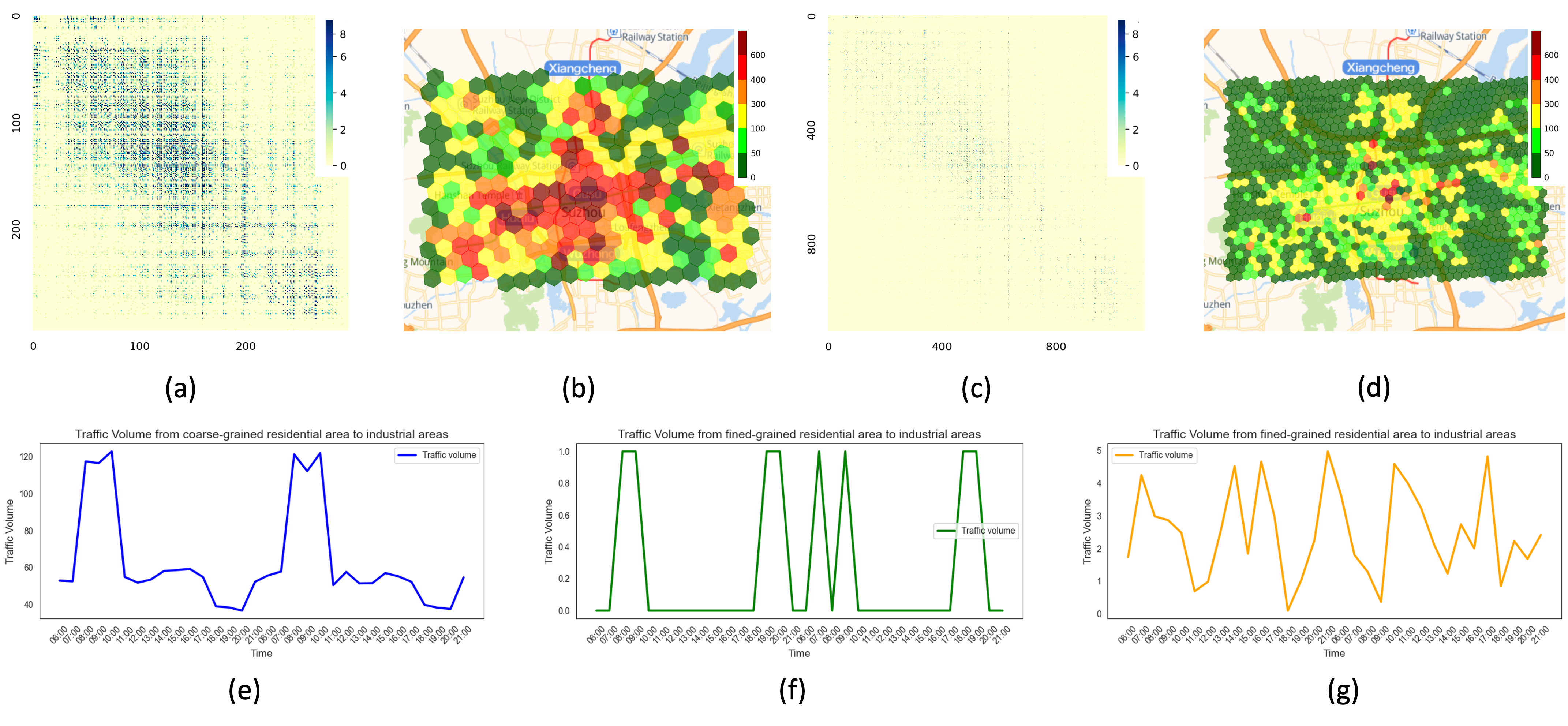} 

\caption{(a) OD matrix of $N=297$ selected cells in City-S; (b) the outflow counts from these cells in (a) within a given day; (c) OD matrix of $N=1111$ selected cells in City-S; and (d) the outflow counts from these cells in (c) within a given day. The temporal trends of an OD flow between residential and industrial areas, depicted at both coarse-grained (e) and fine-grained spatial levels (f) and (g).}
\label{fig:one}
\end{figure*}

These two datasets feature fine-grained spatial resolution, unlike traditional coarse aggregations. Figure \ref{fig:one} (b) and (d) show maps of coarse-grained and fine-grained cells for City-S, with heat maps of daily traffic outflow. Figure \ref{fig:one} (a) and (c) depict the OD matrix of these cells, revealing greater sparsity at the fine-grained level.

Traditionally, OD flows represent large, homogeneous city areas like residential, commercial, or industrial zones. These datasets subdivide large areas into smaller cells, increasing OD data sparsity. For example, Figure \ref{fig:one}(e) shows high traffic volumes from a residential area to industrial areas at the coarse-grained level. When divided into finer cells, as shown in Figure \ref{fig:one}(f) and (g), demand is split, losing temporal periodicity, and sparse traffic demands dominate. The finer-grained spatial division poses significant challenges to traditional OD prediction methods in transportation dynamics. The traffic patterns in these datasets lack the regular periodicity or trends that conventional methods rely on. Therefore, the nuanced characteristics of fine-grained division require novel methodologies for robust learning and prediction in these complex datasets.

The finer-grained spatial division introduces significant challenges to traditional OD prediction methods, particularly in the context of transportation dynamics. The observed traffic patterns in these datasets violate the regular periodicity or trends that conventional statistical or deep-learning methods often rely on. As such, the nuanced characteristics identified through fine-grained spatial division necessitate the exploration of novel methodologies capable of robust learning and prediction under the complexity of these datasets.

In particular, the new method needs to address three  critical questions:
(Q1) How can we reduce prediction errors for OD flows with typically non-zero values, given that the model often yields sparse predictions due to the prevalence of zero values across most spatio-temporal locations?
(Q2) How can we accurately predict exceptional or extreme non-zero demands, which may arise when fine-grained spatial divisions result in less evident periodic patterns?
(Q3) How can we utilize both semantic and geographic information to better understand traffic patterns of OD demands and improve prediction performance?
These questions drive the methodological development outlined in Section \ref{sec3}.

\section{Spatio-temporal prediction model}\label{sec3}
\begin{figure*}[!hbt]
\centering
\includegraphics[width=\textwidth]{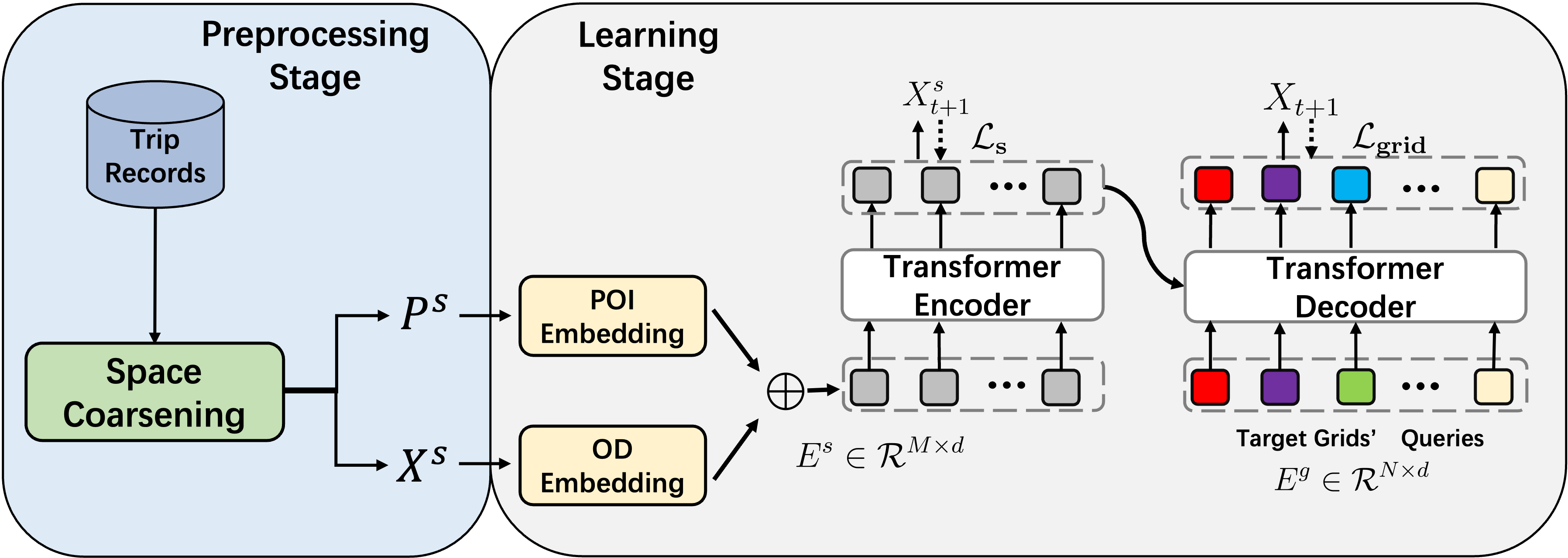}    
\caption{The architecture of OD-CED, consists of a \textit{Preprocessing Stage} and a \textit{Learning Stage}. In the Preprocessing Stage, a space coarsening procedure is applied to adaptively merge fine-grained cells into coarse-grained super-cells. In the Learning Stage, an encoder-decoder network is employed for coarse-grained encoding and fine-grained decoding.}
\label{framework}
\end{figure*}

In this section, we present the detailed architecture of the OD-CED model.
As shown in Figure \ref{framework}, OD-CED comprises two main stages: the \textit{preprocessing stage} and the \textit{learning stage}. In the preprocessing stage, we propose a novel non-parametric down-sampling method to perform space coarsening, merging the original $N$ fine-grained cells into $M$ coarse-grained super-cells to mitigate sparsity issues. In the learning stage, we introduce a novel OD embedding module to capture both the in-flow and out-flow information of super-cells. We design a specialized cross-shaped receptive field to precisely quantify the semantic relationships among super-cells. The encoder takes the concatenation of the learned OD embedding and some external POI embedding as input to refine the representations of super-cells, which are fed into the decoder along with a trainable embedding table of cells.

\subsection{Problem statement}
We first introduce some important notations and formally define the prediction problem.

$\bullet$ \textbf{\textit{Cells.}} We divide the entire urban area into $N$ non-overlapping hexagonal local regions, referred to as "cells", denoted by $G = \left\{g_{1}, g_{2}, \ldots, g_{N} \right\}$. The coverage area of each cell is defined by its center's longitude, latitude, and cell radius. 

$\bullet$ \textbf{\textit{Super-cells.}} Super-cells are sub-regions formed by merging multiple cells together, where the resulting regions have arbitrary shapes and share no common cells. The set of super-cells obtained from $G$ is denoted as $S = \left\{s_{1}, s_{2}, \ldots, s_{M} \right\}$, where $M << N$. We propose a novel space coarsening method to perform the merging of cells to form super-cells.

$\bullet$ \textbf{\textit{OD Matrix.}} The cell-level OD Matrix at time slot $t$ is defined as $X_t=(x_{t}(i,j)) \in \mathcal{R}^{N \times N}$, where $x_{t}(i,j)\in \mathbb{R}$ represents the total traffic demand from cell $i$ to cell $j$. Similarly, the OD Matrix at the super-cell level is defined as $X_t^s \in \mathcal{R}^{M\times M}$, representing the traffic demand among the $M$ super-cells at time slot $t$. For convenience, we define the series of OD Matrices between time slot $t_1$ and $t_2$ as an OD tensor $X_{t_1, t_2} = [ X_{t_1}, X_{t_1+1}, \ldots, X_{t_2} ] \in \mathbb{R}^{N \times N \times (t_2-t_1)} $, and similarly, we let $X^s_{t_1, t_2} = [ X^s_{t_1}, X^s_{t_1+1},  \ldots,  X^s_{t_2}]\in \mathbb{R}^{M \times M \times (t_2-t_1)} $ be the corresponding super-cell level OD tensor.

$\bullet$ \textbf{\textit{POI Matrix.}} We categorize points of interest (POIs) into $p$ different categories and represent the POI information at the super-cell level using a binary matrix $P^s=(p^s(i,j)) \in \mathcal{R}^{M\times p}$. The matrix element $p^s(i,j)$ is equal to 1 if a POI of category $j$ (such as a restaurant or a cinema) is present in super-cell $i$, and 0 otherwise.

$\bullet$ \textbf{\textit{Sparsity rate of OD Matrix.}} The sparsity rate of an origin-destination (OD) tensor $X_{t_1, t_2} \in \mathcal{R}^{N \times N \times (t_2 - t_1)}$ is the proportion of zero elements in the tensor, given by:
\begin{equation}
S = \frac{\sum_{i=1}^N\sum_{j=1}^N \sum_{t=1}^{t_2-t_1} \mathbb{I}(x_{i,j,t}=0)} {N^2(t_2-t_1)},
\end{equation}
where $\mathbb{I}(x_{i,j,t}=0)$ is an indicator function that returns 1 if $x_{i,j,t}$ is zero, and 0 otherwise.

The OD prediction problem involves forecasting the future OD demands $X_{t+1, t+\tau}$ for the next $\tau$ time slots time slots based on past OD data $X_{t-K+1,t}$. 

\subsection{Zero-Inflated Negative Binomial (ZINB)
Distribution}\label{sec3.2}

In this study, we formulate the OD prediction problem in this paper as a non-parametric estimation problem of zero-inflated negative binomial (ZINB) distribution, as inspired by \cite{zhuang2022uncertainty}.
As discussed in Section ~\ref{sec2}, Fine-grained OD data are often characterized by significant sparsity, manifesting as an overabundance of zero counts. The ZINB distribution effectively addresses zero inflation by integrating a new parameter with an NB distribution. This integration allows the model to account for the excess zeros while capturing the variability in the count data. Formally, we assume that for each $t$ and all $t+1 \leq s\leq t+\tau$, the conditional distributions of $x_s(i,j)$'s given historical OD tensor $X_{t-K+1,t}$ are independent and each $x_s(i,j)$ follows a ZINB distribution
\begin{equation}
\small
f_{ZINB}(x_s(i,j)|X_{t-K+1,t})= \left\{\begin{matrix}
 \pi_{s}(i,j) + (1-\pi_{s}(i,j))f_{NB}(0; n_{s}(i,j), p_{s}(i,j)),   &~~~ \text{if} & x_{s}(i,j) = 0; \\
 (1- \pi_{s}(i,j))f_{NB}(x_{s}(i,j); n_{s}(i,j), p_{s}(i,j)),   &~~~ \text{if} & x_{s}(i,j) >  0, 
\end{matrix}\right.
\end{equation}

Here, $f_{NB}$ denotes the negative binomial distribution characterized by shape parameters $n_{s}(i,j)$ and $p_{s}(i,j)$, while $\pi_{s}(i,j)$ represents the probability of an excess zero occurring. The parameters $\pi_s(i,j)$, $n_s(i,j)$, and $p_s(i,j)$ are modeled as functions of the historical OD tensor $X_{t-K+1,t}$ and are estimated using neural networks. This parameterization enables the model to learn complex, non-linear relationships within the data, thereby enhancing its capacity to handle the sparsity inherent in OD datasets.

\subsection{Prepossessing}\label{Space Coarsening}
The preprocessing module comprises four main steps: labeling, construction of transition matrices, propagation, and cell merging.
The motivation behind this module is rooted in the observation that only a few cells exhibit high traffic volumes for most of the time, acting as "dense cells". Meanwhile, the majority of cells have extremely sparse traffic demands, interacting with only a small fraction of these dense cells.
Thus, each dense cell can be considered the center of a community with the "sparse cells" in the same community being the most related ones. After space coarsening, we group each dense cell with nearby sparse cells that share close traffic relationships, forming the upper-level super-cell.

$\bullet$ \textbf{\textit{Labeling.}} Given the historical OD data from time $t=1$ to time $\Tcal_0$,
we calculate a flow statistics $f_i = {\Tcal_0}^{-1}\sum_{t=1}^{\Tcal_0}\left[\sum_{j=1}^N x_{i,j,t}+ \sum_{k=1}^N x_{k,i,t} \right]$ for each cell $g_i$,  which is the mean value of OD flows associated with $g_i$ over the $\Tcal_0$ time slots. 
We can obtain $F'=[f_{(1)},f_{(2)}, \ldots , f_{(N)}]$ satisfying $f_{(1)} \leq f_{(2)} \leq \ldots  \leq f_{(N)}$ by re-ordering the $N$ elements in $F=[f_1,f_2, \ldots,  f_N]$. Let $g_{(i)}$ be the cell corresponding to $f_{(i)}$ for $i=1, \ldots, N$. 
Based on $F'$, the $N$ cells can be divided into two parts: the first $N-M$ cells of $F'$, which have near-zero volumes for all the in- and out-flows, form a sparse set $G_{s} = \left\{ g_{(N-M)}, \ldots , g_{(N)} \right\}$, and the remaining $M$ cells form a dense set $G_d= \left\{ g_{(1)}, ..., g_{(M)} \right\}$, which includes all the dense cells.

We first assign a unique label to each of the $M$ dense cells in $G_d$ such that
any two of them would not fall into the same community. The corresponding initialized labeling matrix of $G_d$ is denoted by
$Y_{l} = \left[ y_{1}, y_{2}, \ldots, y_{M} \right] = \mathbf{I}_{M \times M} \in \mathcal{R}^{M \times M}$, where each $y_{i}$ is an one-hot labelling vector for $g_i$ with only the $i$-th element being 1.  
Since all the $N-M$ "sparse cells" in $G_{s}$ are unassigned to any community at the beginning, we initialize their labeling matrix as a zero matrix $Y_{u} = \mathbf{0}_{(N-M) \times M} \in \mathcal{R}^{(N-M) \times M}$. Thus, the initial labeling matrix $Y^0$ for all the $N$ cells is the concatenation of $Y_{l}$ and $Y_{u}$, denoted by:

\begin{equation}
Y^0 = [ Y_l^T ~~ Y_u^T ]^T
=[\mathbf{I}^T ~~ \mathbf{0}^T]^T \in \mathcal{R}^{N \times M}. 
\end{equation}

$\bullet$ \textbf{\textit{Transition Matrices Construction.}} We introduce two transition matrices that represent the probability of both cells $j$ and   $i$ belonging to the same community from two different perspectives.
One is a semantic transition matrix $T^{sem} \in \mathcal{R}^{N \times N}$ for measuring the traffic connections among cells. Cells connected by large OD flows usually tend to share similar traffic patterns, and are more likely to be categorized into the same community. 

The other is a geographic transition matrix $T^{geo}$ for capturing the geographic adjacency. Based on the first law that everything is related to everything else but nearby things are more related than distant things \citep{tobler1970computer, zhu2018spatial}, geographically adjacent cells usually share certain kinds of common traffic patterns and it is reasonable to merge them together.  
Thus, from the geographic perspective, the probability of $g_j$ belonging to the same community with $g_i$ is defined as 
$ 
    T_{i, j}^{geo} =  {\mathbb{I}(dis(g_{i}, g_{j})<l)}/\{\sum_{j=1}^{m}\mathbb{I}(dis(g_{i}, g_{j})<l)\}, 
$ 
where $dis$ denotes the geographic distance between the centers of two cells and $l$ is a pre-defined threshold. In practice, we let $l$ be slightly larger than twice the cell radius to make $T_{i, j}^{geo}$ between $g_{i}$ and each of its six neighboring cells be 1/6 and others be 0.

This design ensures that each community contains only one dense cell. Dense cells, representing urban residential, work zones, and entertainment venues in practice, exhibit distinct traffic patterns, not only in volume but also in their phases, with observed periodicity. By avoiding amalgamating these cells into larger communities, which would consist of multiple dense cells, the design preserves the model's ability to discern nuanced traffic patterns within the OD matrix. This configuration is essential for accurately capturing the intricate dynamics of traffic flow within and between these communities.

$\bullet$ \textbf{\textit{Propagation.}} Given the initial labeling matrix $Y^0$ and  $T^{sem}$ and $T^{geo}$ defined above, we infer the labels of sparse cells through a five-step labeling propagation procedure as follows:

Step 1: Set $Y^{0, sem} = Y^0$ and $Y^{0, geo} = Y^0$. 
Step 2: Labelling Update: For the $n$-th iteration, $Y^{n, sem} = T^{sem} Y^{n-1}$ and $Y^{n, geo} = T^{geo} Y^{n-1}$.
Step 3: Label Fusion: $Y ^{n} = \alpha Y^{n, sem} + \beta Y^{n, geo}$ with $\alpha + \beta=1$. 
Step 4: Reset the first $M$ rows of $Y^{n}$ to be an identity matrix. 
Step 5: Repeat Steps 2 to 4 until convergence to obtain the labelling matrix $\tilde Y = [ \tilde Y_l^T \tilde Y_u^T]^T.$ 
Step 6: Apply row-wise Argmax operator to $\tilde Y$ to get the final clustering results $\widehat{Y}$.

Note that in Step 3, we set $\alpha=\beta=0.5$ to ensure that both semantic and geographical transitions equally contribute to the labeling propagation, facilitating the convergence of the algorithm.

In Step 4, to prevent merging dense cells, we reset the first $M$ rows of $Y^n$ to an identity matrix at the end of each iteration. This ensures that each resulting community contains only one dense cell. In Step 6, we apply row-wise Argmax to $\tilde Y$ to obtain $\widehat{Y}=(\widehat{Y}_{i,j})$, where $\widehat{Y}_{i,j} = I(\tilde{Y}_{i,j}= \max_{k \in \{1,\ldots, M\}} \tilde{Y}_{i,k})$. This assigns each sparse cell to the community that is most related to from both semantic and geographical perspectives.

$\bullet$ \textbf{\textit{Cell Merging.}} In this step, all the $N$ cells are divided into $M$ communities and each community represents a super-cell. Based on $\widehat{Y}$, we can finally build the coarsened OD tensor $X^{s}_{0,\Tcal_0} = X_{0,\Tcal_{0}} \times_{1} \widehat{Y} \times_{2} \widehat{Y}  \in \mathcal{R}^{M \times M \times \Tcal_0 }$ of length $\Tcal_0$, where $\times_{i}$ denote $i$-mode product of tensor $X_{0,\Tcal_{0}}$ with matrix $\widehat{Y}$ for $i \in \{1, 2 \}$
\footnote{The $n-$mode~(matrix)~product of a tensor $\mathcal{X} \in \mathcal{R} ^{I_1 \times I_2 \times \ldots \times I_N}$ with a matrix $U \in \mathcal{R} ^{J \times I_n}$ is denoted by$\mathcal{X} \times_{n} U$ and is of the size $I_1 \times \ldots \times I_{n-1} \times J \times I_{n+1} \times \ldots \times I_{N}$. Elementwisely, we have $(\mathcal{X} \times_{n} U)_{i_1 \ldots i_{n-1}ji_{n+1} \ldots i_N} = \sum_{i_n=1}^{I_n}x_{i_1i_2 \ldots i_N}u_{i_nj}$.}. Thus, each flow in $X^{s}_{0,\Tcal_0} $ is the summation of a group of OD flows in $X_{0,\Tcal_0}$, 
 making $X^{s}_{0,\Tcal_0}$ much smaller in terms of the size of tensor, yet much denser than $X_{0,\Tcal_0}$.

Our space coarsening module offers two advantages over traditional downsampling methods. First, it retains the original traffic pattern by reducing dimensions at the cell level, unlike methods like mean or max pooling, which can merge unrelated OD flows and alter the data structure. Second, our method accounts for both geographic and semantic dependencies, whereas traditional approaches often ignore semantic relationships among OD flows.

\begin{figure*}[!hbt]
\centering
\includegraphics[width=\textwidth]{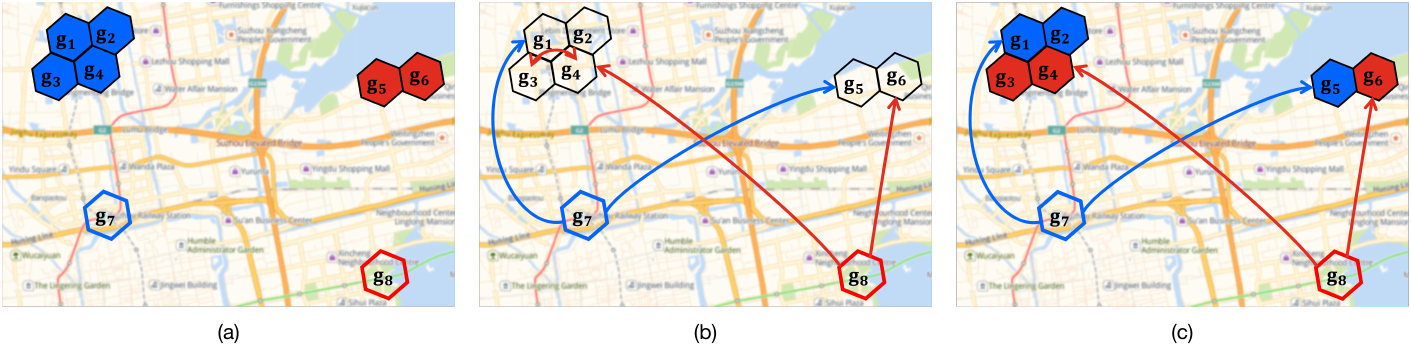}    
\caption{(a)  Cell merging based on geographical closeness. Cells clustered together are assigned the same color; (b) Directed OD flows between cells;  and (c) Clustering result by our method which jointly considers geographic and semantic dependence. }
\label{fig:fig3}
\end{figure*}

For example, in Figure \ref{fig:fig3} (b), $g_1$ and $g_5$ are geographically distant cells, yet they share similar traffic patterns due to a common OD neighbor $g_7$. Conversely, $g_1$ and $g_3$ are close to each other, but their traffic patterns can differ significantly when there are few traffic demands between them. Figure~\ref{fig:fig3} (a) illustrates that some conventional spatial clustering methods based solely on geographic relationships fail to merge $g_1$ with $g_5$, whereas our method successfully groups them together by considering their high semantic dependence.

\subsection{OD Embedding}\label{od_embed}
We present a novel OD embedding module illustrated in Figure \ref{fig:fig4}(b) to learn feature vectors for each super-cell, capturing both traffic flow and Point of Interest (POI) information.
For each super-cell $s_i$, we define one origin matrix $O_{i} = X^{s}_{t-K+1, t}(i,\cdot,\cdot) \in \mathcal{R}^{M \times K}$ and one destination matrix $D_{i} = X^{s}_{t-K+1, t}(\cdot,i,\cdot) \in \mathcal{R}^{M \times K}$, where $O_{i}$ includes all the coarsened OD flows starting from $s_{i}$ and $D_{i}$ includes the ones ending at $s_{i}$. We multiply $O_{i}$ and $D_{i}$ by two weight matrices $W^{o} \in \mathcal{R}^{K \times d}$ and $W^{d} \in \mathcal{R}^{K \times d}$, respectively, to learn temporal representations, leading to two embedded matrices $\widehat{O}_{i} = O_{i} W^{o} \in \mathcal{R}^{M \times d}$ and $\widehat{D}_{i} = D_{i} W^{d} \in \mathcal{R}^{M \times d}$.
Each row of $\widehat{O}_{i}$ (or $\widehat{D}_{i}$) is a $d$-dimensional feature vector, encoding the temporal information of the corresponding coarsened OD flows related to $s_i$.

Let $\widehat{o}_{r}^{i} \in \mathcal{R}^d$ be the $r$-th row of $\widehat{O}_{i}$, $\widehat{d}_{r}^{i} \in \mathcal{R}^d$ be the $r$-th row of $\widehat{D}_{i}$, and $Q=[q_1,\ldots, q_{N_q}]^{\intercal} \in \mathcal{R}^{N_q \times d}$ be a set of $N_q$ randomly initialized learnable queries. 
We define the similarity score between $\widehat{o}_{r}^{i}$ and $q_j$  and that between  $\widehat{d}_{r}^{i}$ and $q_j$ as 
\begin{equation}\label{sim1}
    \alpha^{\widehat{o}, i}_{r,j} =  \frac{\exp(\widehat{o}_{r}^{i \intercal} \cdot {q_{j}})}{\sum_{k=1}^{N_q} \exp(\widehat{o}_{r}^{i \intercal} \cdot {q_{k}})} ~~\mbox{and}~~
    \alpha^{\widehat{d}, i}_{r,j} =  \frac{\exp(\widehat{d}_{r}^{i \intercal} \cdot {q_{j}})}{\sum_{k=1}^{N_q} \exp(\widehat{d}_{r}^{i \intercal} \cdot {q_{k}})},
\end{equation}
respectively. 
A larger similarity score represents higher importance of $\widehat o_r^i$ or $\widehat d_r^i$ to the super-cell $s_i$. The final traffic flow representation of $s_i$ is defined as 
\begin{equation}\label{embed}
   e_{od, i}^{s} = \sum_{r=1}^{M}\sum_{j=1}^{N_{q}} \alpha^{\widehat{o}, i}_{r,j} \cdot \widehat{o}_{r}^{i} + \sum_{r=1}^{M}\sum_{j=1}^{N_{q}} \alpha^{\widehat{d}, i}_{r,j}  \cdot \widehat{d}_{r}^{i},  
\end{equation}
which takes all the coarsened OD flows related to $s_i$ into consideration. By applying this aggregation process to all the $M$ super-cells, we can finally obtain a traffic feature matrix $E^s_{od} = [e^s_{od,1}, e^s_{od,2}, \ldots, e^s_{od,M}]^T \in \Rcal^{M\times d}$.

This design offers two main advantages. Firstly, the architecture entails a fixed number of learning parameters. However, employing a simple aggregation method like a weighted sum of all the related OD flows, such as $e_{od, i}^{s} = \sum^{M}_{r=1} w_{r}^{o}  \widehat{o}_{r}^{i} + \sum^{M}_{r=1} w_{r}^{d} \widehat{d}_{r}^{i}$

requires $2M$ parameters. This leads to a linear increase in parameters with the number of cells, resulting in computational inefficiency as the dataset grows.

Secondly, the aggregation process is permutation invariant such that the learned representation of each super-cell depends solely on the traffic demands of all related OD flows and remains unchanged regardless of their order in the input matrices $\widehat{d}_{r}$ and $\widehat{o}_{r}$. Thus, re-ordering the $M$ rows of the coarsened OD matrix does not change the value of the learned feature vector when $w_{r}^{o}$ and $w_{r}^{d}$ are provided. 
However, for the naive weighted sum, if we exchange any two rows of $\widehat{O}_i$ and $\widehat{d}_i$, then the output $e^s_{od,i}$ changes accordingly, which is unreasonable.

To incorporate POI information, we multiply the super-cell level POI matrix $P^s \in R^{M \times p}$ by a weight matrix $W_{poi} \in R^{p \times d}$ to get the POI representation $E^s_{poi} = P^sW_{poi} \in R^{M \times d}$, and the final OD embedding for each super-cell is computed as $E^s=E^s_{od}+E^s_{poi} \in R^{M\times d}$.

\subsection{OD Encoder-Decoder}

\begin{figure*}[!hbt]
\centering
\includegraphics[width=\textwidth]{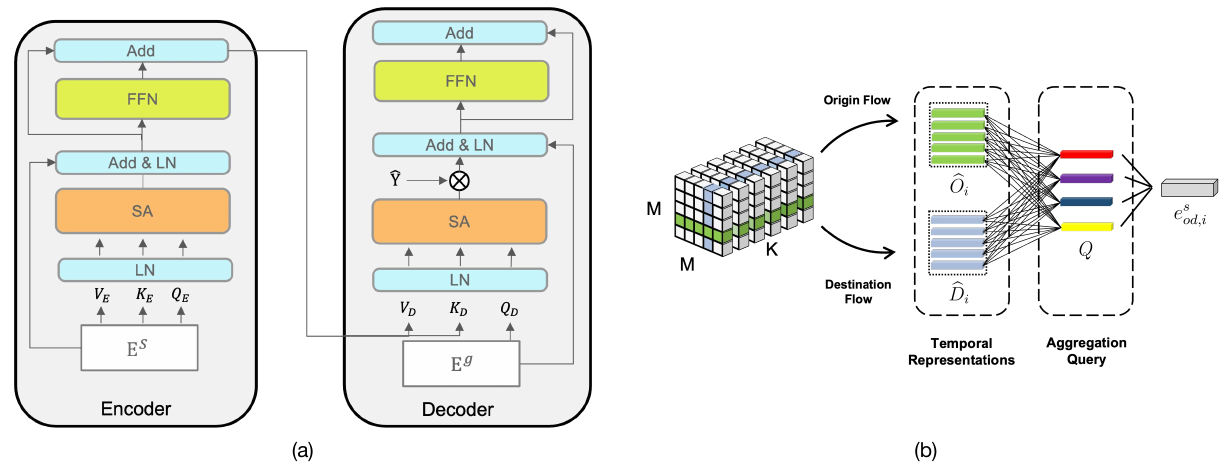}    
\caption{(a) The main architecture of the OD Encoder-Decoder module and (b) The architecture of the OD Embedding module.}
\label{fig:fig4}
\end{figure*}

The OD Encoder-Decoder, depicted in Figure \ref{fig:fig4}(a), serves as the core learning module of the proposed method. Utilizing the OD embedding $E^{s}$, the encoder discerns geographical dependencies among super-cells, and its output is subsequently fed into the decoder to derive representations of fine-grained cells.

\textbf{OD Encoder}. The core of the OD Encoder utilizes a multi-head self-attention (MHSA) architecture \citep{vaswani2017attention}. Each head assesses spatial-temporal similarities among super-cells and refines their representations accordingly. By integrating information from diverse representation subspaces, the multi-head design enhances expressive capabilities, leveraging the low-rank and sparse nature of individual heads \citep{wang2020linformer}.

We assume that all the $h$ heads share the same architecture,  but allow them to have different parameters. 
For each head $i \in  \{1, \ldots, h\}$, the input $E^s$ is projected by three weight matrices $W^{enc}_{Q,i} \in \mathcal{R}^{d \times d_Q}$, $W^{enc}_{K,i} \in \mathcal{R}^{d \times d_K}$, and $W^{enc}_{V,i} \in \mathcal{R}^{d \times d_V}$, respectively, to get 
$Q_{i,E} = E^{s} W^{enc}_{Q,i}$, $K_{i,E} = E^{s} W^{enc}_{K,i}$, and  $V_{i,E} = E^{s} W^{enc}_{Q,i}$,
where $d_Q=d_V=d_K=\lfloor \frac{d}{h} \rfloor$. 
Then, the output of the $i$-th head $H_i$ is given by $H^{enc}_i= softmax\left( \frac{Q_{i,E} K_{i,E}^T}{\sqrt{d}}\right)V_{i,E}$. 

In this case, the $j$-th row of $H^{enc}_i$ represents a weighted sum of the rows in $V_{i,E}$, where the weights correspond to the similarities between the $j$-th row of $Q_{i,E}$ and the rows of $K_{i,E}$. Subsequently, we multiply the concatenation of the $h$ heads by a weight matrix $W^{enc}_O \in \Rcal^{h \lfloor \frac{d}{h} \rfloor\times d}$ to obtain the output $MHSA(E^s)= [ H_1^{enc}, \ldots, H_h^{enc} ]W^O$ of this MHSA module.

Specifically, we apply layer normalization (LN) to the input $E^{s}$ and utilize a residual block design to enhance optimization efficiency \citep{narang2021transformer}.
Therefore, the final architecture of the MHSA module is as $E^{s \prime} = MHSA(LN(E^{s})) + E^{s}$
and $E^{s \prime}$ is then fed into a feed-forward network (FFN) to get the final output $\widehat{E}^{s} = FFN(LN(E^{s\prime})) + E^{s\prime}$. 
The FFN  here is simply a two-layer fully connected network with ReLU. 
$\widehat{E}^{s} = \left \{ \widehat{e}^{s}_{1}, \widehat{e}^{s}_{2}, \ldots, \widehat{e}^{s}_{M} \right \} $ here encodes pairwise semantic relationship between super-girds. 

\textbf{OD Decoder}. The decoder utilizes a randomly initialized trainable embedding matrix $E^{g} \in \mathcal{R}^{N \times d}$ along with the output of the OD encoder $\widehat{E}^{s}$ to learn the representations of fine-grained cells. Incorporating $E^{g}$ enables the OD-CED model to capture the long-term characteristics of fine-grained cells through end-to-end training with the entire dataset. The decoder employs a Multi-Head-Cross-Attention (MHCA) module, which measures spatial-temporal similarities between super-cells and fine-grained cells, aggregating the embedding of related super-cells to obtain the representation of the target fine-grained cell.

For each head $i \in \{1,\ldots, h\}$, we project $E^{g}$ with $W_{Q,i}^{dec} \in\mathcal{R}^{d \times d_Q}$ to obtain the embedding $Q_{i,D} = E^gW_{Q,i}^{dec}$ of fine-grained cells. Similarly, we project $\widehat{E}^{s}$ with $W_{V,i}^{dec} \in \mathcal{R}^{d \times d_V} $ and $W_{K,i}^{dec} \in \mathcal{R}^{d \times d_K}$ to obtain the embeddings $V_{i,D} = \widehat{E}^sW_{V,i}^{dec}$ and $K_{i,D} = \widehat{E}^sW_{K,i}^{dec}$ of super-cells, where $d_Q=d_V=d_K=\lfloor \frac{d}{h} \rfloor$. We then re-weight $V_{i,D}$ using the similarity between 
$K_{i,D}$ and 
$Q_{i,D}$ to derive the output of the $i$-th head $H^{dec}_i = \left[\widehat{Y} \otimes softmax\left( \frac{Q_{i,D} K_{i,D}^T}{\sqrt{d}}\right)\right]V_{i,D}$.

$softmax(Q_{i,D} K_{i,D}^T/\sqrt{d})\in \Rcal^{N\times M}$ computes the similarity between the features of the $N$ fine-grained cells and those of the $M$ super-cells. The $\widehat{Y} \in \{0,1\}^{N \times M}$ from the Propagation step indicates whether a fine-grained cell belongs to the same community as a super-cell, and its dot product with the softmax term helps to downplay weak connections among cells. Consequently, the representations of each fine-grained cell are influenced by its neighbors within the same community.

We then multiply the concatenation of the $h$ heads with a weight matrix $W^{dec}_O \in \Rcal^{h \lfloor \frac{d}{h} \rfloor\times d}$ to get the output $MHCA(E^g,\widehat{E}^s) = [ H^{dec}_1, \ldots, H^{dec}_h ]W^{dec}_O$.

The residual mechanism, along with LN is applied, and the final representation $\widehat{E}^g$ of each fine-grained cell $g$ is obtained as follows,
\begin{equation}
 E^{g \prime} = MHCA(LN(E^g),LN(\widehat{E}^s)) + E^{g}~~~\mbox{and}~~~
 \widehat{E}^{g} = FFN(LN(E^{g\prime})) + LN(E^{g\prime}).
\end{equation}
The complexity of the OD Decoder is $\Omega(N \times M) \approx \Omega(N)$, as $M$ is much smaller than $N$ in practice and can be ignored. This makes our OD-CED model more scalable and computationally efficient compared to most existing OD prediction methods. Moreover, by introducing the cell embedding table $E^{g}$, we eliminate direct computations at the original fine-grained level, further reducing the computational complexity.

\subsection{Optimization and Prediction} 
Using the spatial-temporal representations of fine-grained cells, denoted as $\widehat{E}^{g} \in \mathcal{R}^{N \times d}$, and super-cells, represented as $\widehat{E}^{s} \in \mathcal{R}^{M \times d}$ as input, 
we employ a neural network with linear transformations and $1 \times 1$ convolutions to predict $\{n_s(i,j), p_s(i,j), \pi_s(i,j); t+1 \leq s\leq t+\tau, 1 \leq i,j\leq N\}$ of the ZINB distribution as defined in Section \ref{sec3.2}. The final optimization of the full OD-CED model parameterized with $\theta$ is to minimize the negative log-likelihood $-\sum_{t=1}^T \log(L(X_{t+1,t+\tau}|X_{t-K+1, t};\theta))$, where $L(X_{t+1,t+\tau}|X_{t-K+1,t})$ is the conditional likelihood function of $X_{t+1,t+\tau}$ given $X_{t-K+1,t}$ defined as follows,
\begin{equation}
    L(X_{t+1,t+\tau}|X_{t-K+1,t}) = \prod_{s=1}^{\tau}\prod_{i = 1}^N\prod_{j=1}^N f_{ZINB}(x_s(i,j); n_s(i,j), p_s(i,j), \pi_s(i,j)).
    \label{eq:likelihood}
\end{equation}

Upon obtaining $\hat\theta$, we naturally obtain estimators for $\hat{n}_s(i,j)$, $\hat{p}_s(i,j)$, and $\hat{\pi}_s(i,j)$. Subsequently, we utilize these estimators to predict the future OD tensor $X_{t'+1,t'+\tau} \in \mathcal{R}^{N \times N \times \tau}$ based on $X_{t'-K,t'}$, leveraging the expectation of the conditional mean as follows,

\begin{equation}
    \hat{x}_s(i,j) = (\frac{1}{\hat{p}_s(i,j)}-1)\hat{n}_s(i,j) ~~\mbox{for}~~~  t'+1 \leq s \leq t'+\tau~~\mbox{and} ~~ 1 \leq i,j\leq N. 
\end{equation}

\section{Real Analysis}
\label{sec4}

In this section, we conduct comprehensive experiments on two fine-grained OD datasets City-C and City-S, as described in Section 2, to evaluate the proposed method. To better illustrate the advantages of our approach in addressing the challenges mentioned in the introduction, we compare OD-CED with SOTA OD prediction models, including two traditional statistical methods Historical Average (HA) and Linear Regression and four deep learning-based methods: CSTN \citep{liu2019contextualized}, MRSTN \citep{noursalehi2021dynamic}, STGCN \citep{song2020spatial}, and GEML \citep{wang2019origin}.

HA predicts future demand by averaging historical demands of the same time point from the most recent days. For the linear regression method, we implement two classic models: Ordinary Least Squares (OLS) Regression and LASSO \citep{tibshirani1996regression}. 
For deep learning models, we used the Adam optimizer, and the maximum number of training epochs was set to 100. We set the total number of super-cells to $M = 60$, approximately one-tenth the size of the total fine-grained cells for both cities. The embedding dimension in Section \ref{od_embed} was set to $d = 64$, and the number of aggregation queries to $Q_{n} = 32$. The initial learning rate was 0.004, halved every 50 epochs, with a fixed batch size of 32.

For both datasets, we collect comprehensive Point of Interest (POI) data, covering 23 categories and 532 subcategories, including weather conditions, location functionalities, and day information. All these contextual features are encoded using One-Hot encoding. 

We evaluated the prediction accuracy of each method using three widely-used metrics: Root Mean Square Error (RMSE), Weighted Mean Absolute Percentage Error (wMAPE), and Common Part of Commuters (CPC). 

We adhere to the approach outlined in prior studies \citep{liu2019contextualized, yao2018deep} for computing the three metrics using non-zero OD demands. This choice is motivated by the fact that flows with no traveling records hold no significance in the ridesharing industry. 

\begin{table}[!hbt]
\begin{center}
\renewcommand\arraystretch{1.3}
\caption{Prediction performance of all the compared methods on the test data across City-C and City-S, evaluated by Root Mean Square Error (RMSE), Weighted Mean Absolute Percentage Error (wMAPE), and Common Part of Commuters (CPC).} 
\label{overall_performance}
\setlength\extrarowheight{-6pt}
\begin{tabular}{l|lll|lll}
\hline
\multicolumn{1}{c|}{\multirow{2}{*}{\textbf{Method}}} & \multicolumn{3}{c|}{\textbf{City-C}}                                                       & \multicolumn{3}{c}{\textbf{City-S}}                                                       \\ \cline{2-7} 
\multicolumn{1}{c|}{}                                 & \multicolumn{1}{l|}{\textbf{wMAPE}}   & \multicolumn{1}{l|}{\textbf{RMSE}}    & \textbf{CPC}     & \multicolumn{1}{l|}{\textbf{wMAPE}}   & \multicolumn{1}{l|}{\textbf{RMSE}}    & \textbf{CPC}   \\ \hline \hline
HA                                                    & \multicolumn{1}{l|}{0.813}            & \multicolumn{1}{l|}{1.442}            & 0.348            & \multicolumn{1}{l|}{0.821}            & \multicolumn{1}{l|}{1.435}            & 0.355  \\       
OLSR                                                  & \multicolumn{1}{l|}{0.822}            & \multicolumn{1}{l|}{1.419}            & 0.324            & \multicolumn{1}{l|}{0.816}            & \multicolumn{1}{l|}{1.351}            & 0.333          \\
LASSO                                                 & \multicolumn{1}{l|}{0.807}            & \multicolumn{1}{l|}{1.424}            & 0.359            & \multicolumn{1}{l|}{0.813}            & \multicolumn{1}{l|}{1.349}            & 0.337          \\
CSTN                                                  & \multicolumn{1}{l|}{0.782}            & \multicolumn{1}{l|}{1.370}            & 0.354            & \multicolumn{1}{l|}{0.721}            & \multicolumn{1}{l|}{1.217}            & 0.451          \\
MRSTN                                                 & \multicolumn{1}{l|}{0.788}            & \multicolumn{1}{l|}{1.380}            & 0.351            & \multicolumn{1}{l|}{0.766}            & \multicolumn{1}{l|}{1.253}            & 0.464          \\
GEML                                                  & \multicolumn{1}{l|}{0.667}            & \multicolumn{1}{l|}{1.255}            & 0.540            & \multicolumn{1}{l|}{0.605}            & \multicolumn{1}{l|}{1.146}            & 0.597          \\
STGCN                                                 & \multicolumn{1}{l|}{0.681}            & \multicolumn{1}{l|}{1.337}            & 0.488            & \multicolumn{1}{l|}{0.596}            & \multicolumn{1}{l|}{1.210}            & 0.674          \\ \hline\hline
\textbf{OD-CED}                                       & \multicolumn{1}{l|}{\textbf{0.411}}   & \multicolumn{1}{l|}{\textbf{0.905}}   & \textbf{0.776}   & \multicolumn{1}{l|}{\textbf{0.323}}   & \multicolumn{1}{l|}{\textbf{0.740}}   & \textbf{0.889} \\ \hline

\end{tabular}
\end{center}
\end{table}
Table \ref{overall_performance} presents the prediction results of each compared method.
OD-CED outperforms all other methods in both datasets, with more significant improvement observed in City-S due to its larger data size. The superior performance of OD-CED can be attributed to two key factors. First, the space coarsening module effectively reduces computation costs and addresses data sparsity by grouping fine-grained cells into super-cells. Second, the OD Embedding module and the Encoder-Decoder capture both pairwise semantic relationships between regions and global geographic patterns.
HA consistently performs poorly as it ignores data variations. Although OLSR and LASSO show slight improvements over HA, they fail to effectively capture the spatial characteristics of OD data, which are crucial for accurate OD prediction \citep{wang2019origin}.
CSTN and MRSTN show a significant increase in both wMAPEs and RMSEs, leveraging CNNs to extract spatial features. However, the improvement in the sparser dataset City-C is much less significant compared to City-S. This suggests that traditional CNNs are not well-suited for OD data, as there are no meaningful relationships between spatially nearby OD flows in an OD matrix. 

GEML achieves the second-best performance in most scenarios as it captures semantic and geographical relationships by combining GNN with skip-LSTM. However, despite its complex model architecture and utilizing more than 30 times more parameters than OD-CED, the wMAPE of GEML is still 29.04\% higher than that of OD-CED.

\begin{figure*}[!hbt]
\centering
\includegraphics[width=0.8\textwidth]{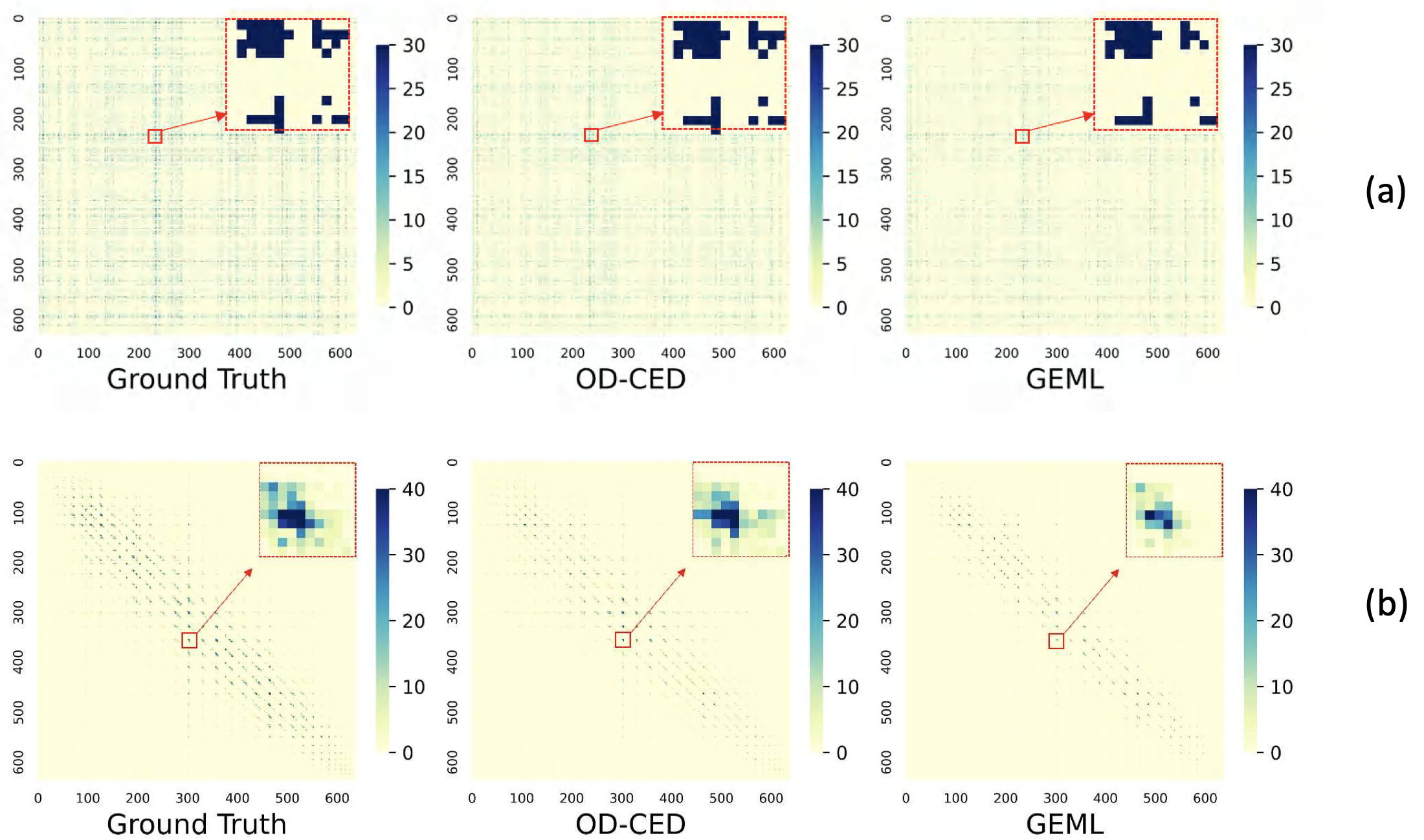}   \caption{(a) The prediction results of the OD-CED and GEML for City-C 
( Nov. 14 ); (b) The prediction results of the OD-CED and GEML for City-S ( Dec. 13 ). }
\label{fig:towcase}
\end{figure*}

To further illustrate why OD-CED outperforms GEML in practice, we conduct a minor case study comparing the prediction results of OD-CED and GEML for two randomly selected days in City-C and City-S. Figures \ref{fig:towcase}(a) and \ref{fig:towcase}(b) depict the ground truths in the first column, followed by the predictions of OD-CED and GEML in the subsequent columns. It's observed that OD-CED exhibits greater sensitivity to extremely large OD demands, leading to an overall more accurate prediction outcome compared to GEML.

To highlight the training efficiency of OD-CED, we compare the training time of each method against its model size. Table \ref{training-time} presents the computation time of each method for one training epoch on a V100 GPU. 
OD-CED is significantly more efficient than the other four methods due to its lightweight design. Even the second-best approach, GEML, requires roughly twice as much training time. Given the necessity for real-time model updates in dynamic platforms, the proposed OD-CED stands out as a practical choice.

We conduct further analysis to demonstrate the OD-CED model's effectiveness in enhancing forecasting accuracy under sparse data conditions. Additionally, we evaluate the significance of each primary module within the OD-CED framework. For more comprehensive details, please refer to the 'Supplementary Results of the Real Data Analysis' section in the supplementary file.

\begin{table}[hbt]
\begin{center}
\renewcommand\arraystretch{1.3}
\caption{Number of Parameters and Training time of one epoch for each compared method} 
\setlength\extrarowheight{-5pt}
\begin{tabular}{l|l|l|l|l|l}
\hline
\textbf{}       & \textbf{CSTN} & \textbf{MRSTN} & \textbf{GEML} & \textbf{STGCN} & \textbf{OD-CED} \\ \hline
\textbf{\# of Params (in millions)} & 0.54M         & 0.67M          & 2.9M         & 1.6M          & \textbf{0.1M}  \\ \hline
\textbf{Training Time (in seconds)}   & 1222.13s      & 1602.14s       & 39.63s        & 28.81s         & \textbf{22.12s} \\ \hline
\end{tabular}
\label{training-time}
\end{center}
\end{table}

\section{Conclusion}
\label{sec: conclusion}
In this paper, we propose a novel prediction model OD-CED for large-scale fine-grained OD data, while addressing three main challenges. We propose a novel space coarsening module to group small fine-grained cells together to highly increase the computation efficiency and preserve the data structure. 
The OD Embedding module captures semantic and geographical dependencies in a global way based on a well-designed permutation invariant operator. 
At last, a novel Encoder-Decoder is introduced to predict the fine-grained OD matrices by utilizing the spatial-temporal information obtained in the coarsened space. 
Some empirical results show that OD-CED can exhibit great improvement over existing methods, especially when the OD data is extremely sparse.

\begin{appendix}
    \section{Descriptions and discussions of the Real data}
    Below is a brief introduction of the two datasets. 
    
    \textbf{City-C:} This anonymized dataset, provided by a ridesharing company, includes historical ride-hailing orders in City C from November 1 to 30, 2016. The city is divided into 632 hexagonal cells, each covering about 1.2 square kilometers, with hourly ride-hailing orders recorded between any two cells. The first two weeks are for model training, with the validation and test sets each spanning one week. The sparsity ratio is 99.15\%, indicating over 99
    
    \textbf{City-S:}. Collected from June 1 to November 30, 2021, in City S, this dataset covers 180 days. The city is divided into 638 segments, each covering 2.1 square kilometers. Hourly orders are recorded, with the last 7 days for testing, and the remaining days equally split for training and validation. The sparsity ratio is 98.02\%, slightly lower than City-C, suggesting higher travel demands.

    These two datasets feature fine-grained spatial resolution, unlike traditional coarse aggregations. Figure \ref{fig:one} (b) and (d) show maps of coarse-grained and fine-grained cells for City-S, with heat maps of daily traffic outflow. Figure \ref{fig:one} (a) and (c) depict the OD matrix of these cells, revealing greater sparsity at the fine-grained level.
    
    Traditionally, OD flows represent large, homogeneous city areas like residential, commercial, or industrial zones. These datasets subdivide large areas into smaller cells, increasing OD data sparsity. For example, Figure \ref{fig:one}(e) shows high traffic volumes from a residential area to industrial areas at the coarse-grained level. When divided into finer cells, as shown in Figure \ref{fig:one}(f) and (g), demand is split, losing temporal periodicity, and sparse traffic demands dominate. The finer-grained spatial division poses significant challenges to traditional OD prediction methods in transportation dynamics. The traffic patterns in these datasets lack the regular periodicity or trends that conventional methods rely on. Therefore, the nuanced characteristics of fine-grained division require novel methodologies for robust learning and prediction in these complex datasets.

    \section{Supplementary results of the Real Data Analysis}
    
    \subsection{Comparison Study}
    In this section, we introduce two more baselines, STZINB-GNN \citep{zhuang2022uncertainty} and STTD \citep{jiang2023uncertainty}, that were recently proposed for the sparse OD prediction problem.
    
    Although, these two methods were proposed to solve the spare problem, neither of them is well-suited for our fine-grained OD data. these approaches model the problem from the perspective of O-D pairs, which results in the computational complexity proportional to the square of the number of spatial cells. Given that our spatial partition scale is significantly finer ($672 \rightarrow 679^2$), these methods are not directly applicable to our problem due to their high computational cost. 
    
    To better assess the effectiveness of our approach compared to STZINB and STTD and achieve a fair comparison, we selected a subset of cells from our dataset and conducted comparative experiments on the corresponding OD matrix. As shown in Table \ref{reply:comparison}, our approach outperforms the alternatives.
    
        \begin{table}[]
        \centering
        \begin{tabular}{l|ccc|ccc}
        \hline
        \multirow{2}{*}{Method} & \multicolumn{3}{c|}{City-C} & \multicolumn{3}{c}{City-S} \\
                                 & wMAPE  & MSE    & CPC    & wMAPE  & MSE    & CPC    \\ \hline
        STZINB                   & 0.7008 & 1.4864 & 0.5559 & 0.6225 & 1.1566 & 0.6047 \\
        STTD                     & 0.7972 & 1.4204 & 0.3372 & 0.7972 & 1.4397 & 0.3592 \\
        \textbf{OD-CED}          & \textbf{0.431}  & \textbf{0.965}  & \textbf{0.792}  & \textbf{0.341}  & \textbf{0.789}  & \textbf{0.92}  \\ \hline
        \end{tabular}
        \caption{Comparison of different methods in City-C and City-S}
        \label{reply:comparison}
        \end{table}

    \subsection{Sparsity Analysis}
\begin{figure*}[!hbt]
    \centering
    \includegraphics[width=0.85\textwidth]{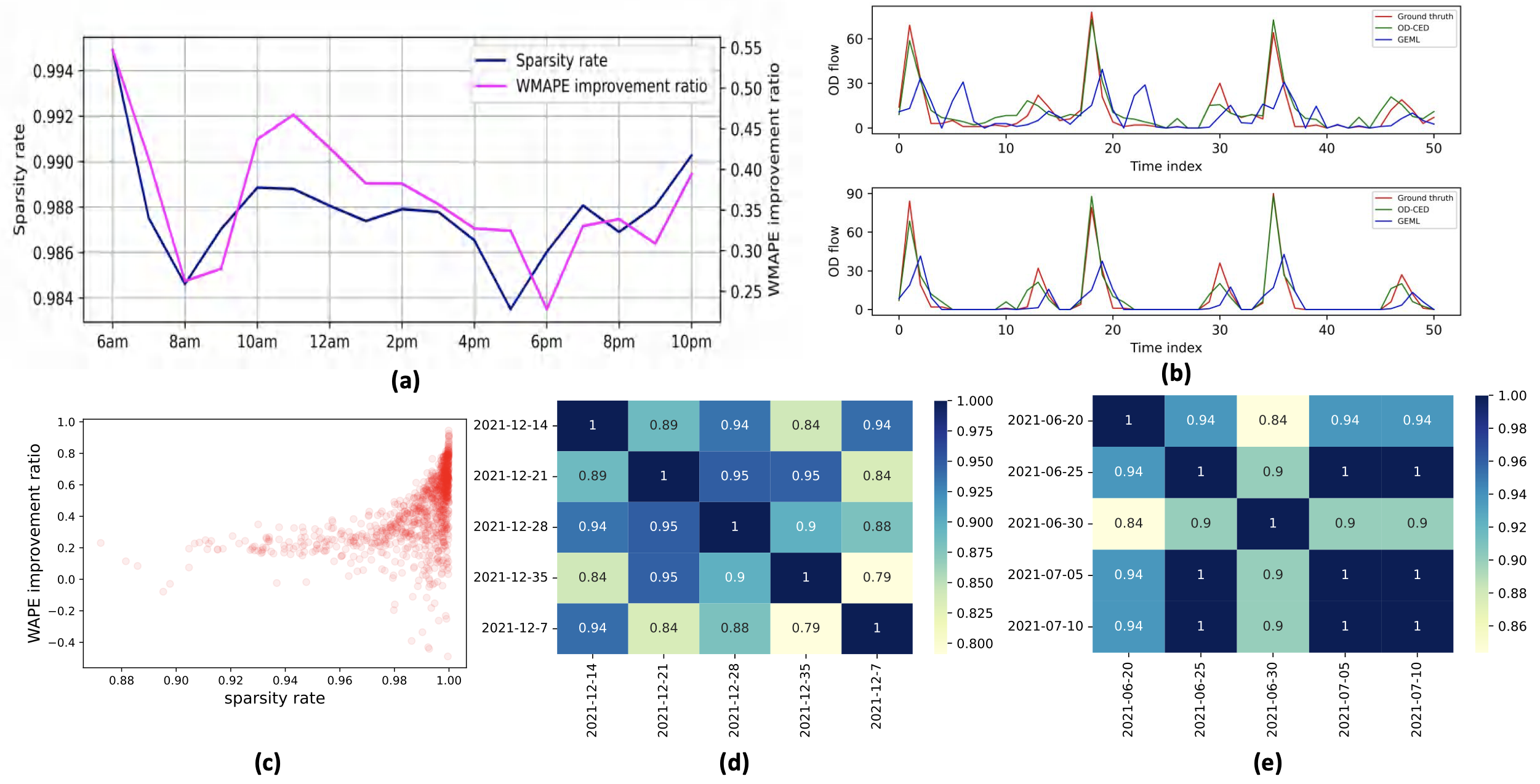}    
    \caption{(a) Temporal dynamics of sparsity rate and wMAPE improvement. (b) The prediction results at two selected OD flows. (c) Comparison between sparsity rate and wMAPE improvement.  Jaccard Similarity between labeling results across different time periods in City-C (d) and  in City-S (e).}
    \label{fig:lastpic}
    \end{figure*}

    In this part, we meticulously assess the OD-CED model's efficacy in improving forecasting accuracy amidst sparse conditions and irregular traffic demand scenarios, using data from the City-S dataset. The main results are primarily illustrated in Figures \ref{fig:lastpic}.
    
    Figure \ref{fig:lastpic} (a) illustrates the temporal dynamics of sparsity rate and the improvement in weighted Mean Absolute Percentage Error (wMAPE) achieved by OD-CED over the baseline GEML. Notably, between 10 a.m. and 4 p.m., we observe a significant reduction in travel demand, leading to heightened data sparsity. This contrasts starkly with the lower sparsity observed during peak hours. Intriguingly, our method exhibits a substantial performance enhancement during off-peak hours, underscoring the OD-CED model's proficiency in navigating the challenges posed by data sparsity.

    Spatial analysis further underscores the superiority of our method. By investigating $1000$ randomly selected $15 \times 15$ sub-regions within the City-C OD tensor, we discern a clear correlation between sparsity rates and the wMAPE improvement ratio of OD-CED over GEML. This observation confirms the strategic advantage of our space coarsening module in consolidating sparse cells to amplify signal strength, as depicted in Figure \ref{fig:lastpic}(c).
    
    From a practical standpoint, the prediction outcomes generated by our method are invaluable for optimizing operational expenditures throughout the day. As illustrated in Figure \ref{fig:lastpic}(b), a comparative analysis of GEML and OD-CED predictions against ground truths for select OD flows during peak periods showcases the OD-CED model's proficiency in capturing daily traffic patterns and delivering precise demand forecasts. This precision is attributed to the OD embedding module's holistic approach to learning semantic dependencies, in stark contrast to GEML's conservative estimations, which are hindered by the prevalent zero entries in the dataset.
    
    \subsection{Ablation Study} 
    To evaluate the necessity of each main module in OD-CED, we conduct an ablation study comparing five different variants: (i) \textbf{w/o Agg-Query}, removing the learnable aggregation query in the OD Embedding module and using a naive summation instead; (ii) \textbf{w/o O-flow}, learning the embedding only from the destination flow matrix without using the origin flow matrix; (iii) \textbf{Linear model}, replacing the OD Embedding part with a simple linear model; (iv) \textbf{w/o Encoder}, removing the Encoder and directly feeding the cells' embedding and POI features into the Decoder; and (v) \textbf{w/o Attention Mask}, removing the Attention mask from the Decoder.

    \begin{table}[hbt]
    \begin{center}
    \renewcommand\arraystretch{1.3}
    \caption{Prediction performance of different variants of OD-CED on the test data across City-C and City-S, evaluated by Root Mean Square Error (RMSE), Weighted Mean Absolute Percentage Error (wMAPE), and Common Part of Commuters (CPC).} 
    \label{ablation study}
    \setlength\extrarowheight{-6pt}
    \begin{tabular}{l|lll|lll}
    \hline
    \multicolumn{1}{c|}{\multirow{2}{*}{\textbf{Method}}} & \multicolumn{3}{c|}{\textbf{City-C}}                                                 & \multicolumn{3}{c}{\textbf{City-S}}                                                   \\ \cline{2-7} 
    \multicolumn{1}{c|}{}                                 & \multicolumn{1}{c|}{\textbf{wMAPE}} & \multicolumn{1}{c|}{\textbf{RMSE}}  & \textbf{CPC}   & \multicolumn{1}{c|}{\textbf{wMAPE}} & \multicolumn{1}{c|}{\textbf{RMSE}}  & \textbf{CPC}   \\ \hline\hline
    w/o Agg-Query                                         & \multicolumn{1}{l|}{0.646}          & \multicolumn{1}{l|}{1.203}          & 0.572          & \multicolumn{1}{l|}{0.519}          & \multicolumn{1}{l|}{1.055}          & 0.681          \\
    w/o O-flow                                            & \multicolumn{1}{l|}{0.645}          & \multicolumn{1}{l|}{1.201}          & 0.561          & \multicolumn{1}{l|}{0.618}          & \multicolumn{1}{l|}{1.212}          & 0.563          \\ \hline
    w/o Encoder                                           & \multicolumn{1}{l|}{0.652}          & \multicolumn{1}{l|}{1.213}          & 0.559          & \multicolumn{1}{l|}{0.554}          & \multicolumn{1}{l|}{1.108}          & 0.634          \\
    w/o Attention Mask                                    & \multicolumn{1}{l|}{0.616}          & \multicolumn{1}{l|}{1.149}          & 0.583          & \multicolumn{1}{l|}{0.517}          & \multicolumn{1}{l|}{1.077}          & 0.703          \\ \hline
    OD-CED (Linear model)                                     & \multicolumn{1}{l|}{0.674}          & \multicolumn{1}{l|}{1.225}          & 0.519          & \multicolumn{1}{l|}{0.591}          & \multicolumn{1}{l|}{1.131}          & 0.604          \\ \hline\hline
    \textbf{Full OD-CED}                                       & \multicolumn{1}{l|}{\textbf{0.411}} & \multicolumn{1}{l|}{\textbf{0.905}} & \textbf{0.776} & \multicolumn{1}{l|}{\textbf{0.323}} & \multicolumn{1}{l|}{\textbf{0.740}} & \textbf{0.889} \\ \hline
    \end{tabular}
    \end{center}
    \end{table}
    
    As shown in Table \ref{ablation study}, the full OD-CED model outperforms all its five variants. OD-CED (Linear model) exhibits nearly the worst performance in both datasets, indicating that incorrect parameter sharing and irrelevant information introduction occur when learning spatial features. Replacing the linear model with our proposed OD embedding module significantly improves both wMAPE and RMSE, underscoring the importance of our permutation invariant design. OD-CED (w/o O-flow) performs second worst, highlighting the importance of both origin and destination flow matrices in capturing pairwise semantic relationships. This result also suggests the significance of hierarchically capturing semantic and geographical dependencies in handling large-scale sparse OD data. Although OD-CED (w/o Attention Mask) performs best among the five variants, its wMAPE remains 16.24\% lower than the full OD-CED model in City-S. This suggests that the full OD-CED model may better capture traffic patterns in OD matrices by aggregating features of cells from the same community determined by the space coarsening module.

    \subsection{Study of Space Coarsening}
    In this section, we aim to empirically compare the Space Coarsening module with two classic downsampling methods, \textit{Low spatial level} and \textit{Mean Pooling}, discussed in the Space Coarsening part of the main text. 
    The \textit{Low spatial level} strategy groups every six geographically nearby hexagonal cells together to build a coarser-grained cell, ignoring semantic connections among cells. As a result, each obtained coarser-grained cell is approximately seven times the size of the original fine-grained cell.
    The \textit{Mean Pooling} strategy divides the OD matrix at each time slot into non-overlapping $4 \times 4$ patches. The mean value of the 16 OD flows within each patch represents the corresponding region, resulting in a down-sampled OD matrix that is one-sixteenth of the size of the original \citep{boureau2010theoretical}.

    \begin{table}[!hbt]
        \centering
        \caption{Empirical comparison between Space Coarsening and two classic downsampling methods in terms of prediction performance the on test data across the two cities, evaluated by Root Mean Square Error (RMSE), Weighted Mean Absolute Percentage Error (wMAPE), and Common Part of Commuters (CPC)}
        \label{Coarsening Strategy Study}
            \begin{tabular}{l|lll|lll}
                \hline
                \multicolumn{1}{c|}{\multirow{2}{*}{\textbf{Method}}} & \multicolumn{3}{c|}{\textbf{City-C}}                                                 & \multicolumn{3}{c}{\textbf{City-S}}                                                   \\ \cline{2-7} 
                \multicolumn{1}{c|}{}                                 & \multicolumn{1}{c|}{\textbf{wMAPE}} & \multicolumn{1}{c|}{\textbf{RMSE}}  & \textbf{CPC}   & \multicolumn{1}{c|}{\textbf{wMAPE}} & \multicolumn{1}{c|}{\textbf{RMSE}}  & \textbf{CPC}   \\ \hline
                Low spatial level                                     & \multicolumn{1}{l|}{0.495}          & \multicolumn{1}{l|}{1.032}          & 0.742         & \multicolumn{1}{l|}{0.510}          & \multicolumn{1}{l|}{0.905}          & 0.801          \\
                Mean Pooling                                          & \multicolumn{1}{l|}{0.524}          & \multicolumn{1}{l|}{1.061}          & 0.717           & \multicolumn{1}{l|}{0.521}          & \multicolumn{1}{l|}{0.914}          &   0.782       \\
                \textbf{OD-CED}                                       & \multicolumn{1}{l|}{\textbf{0.411}} & \multicolumn{1}{l|}{\textbf{0.905}} & \textbf{0.776} & \multicolumn{1}{l|}{\textbf{0.323}} & \multicolumn{1}{l|}{\textbf{0.740}} & \textbf{0.889} \\ \hline
            \end{tabular}
    \end{table}
    
\end{appendix}

\bibliographystyle{agsm}

\bibliography{sample-base}
\end{document}